\pdfoutput=1

\documentclass[final]{cvpr}

\usepackage{times}
\usepackage{epsfig}
\usepackage{amsmath}
\usepackage{amssymb}

\usepackage[ruled,vlined]{algorithm2e}
\usepackage{multirow}
\usepackage{color,soul}
\usepackage{tabularx}
\usepackage{booktabs}

\newcommand{\ignore}[1]{}
\newcommand{\TODO}[1]{{\hl{\textbf{TODO:}}}\xspace}

\usepackage[pagebackref=true,breaklinks=true,colorlinks,bookmarks=false]{hyperref}

\def\cvprPaperID{6437d} 
\def\confYear{CVPR 2021}

\begin{document}

\title{MAZE: Data-Free Model Stealing Attack Using\\Zeroth-Order Gradient Estimation}

\author{Sanjay Kariyappa\\
Georgia Institute of Technology\\
Atlanta GA, USA\\
{\tt\small sanjaykariyappa@gatech.edu}

\and
Atul Prakash\\
University of Michigan\\
Ann Arbor MI, USA\\
{\tt\small aprakash@umich.edu}

\and
Moinuddin K Qureshi\\
Georgia Institute of Technology\\
Atlanta GA, USA\\
{\tt\small moin@gatech.edu}
}

\maketitle

\begin{abstract}
High quality Machine Learning (ML) models are often considered valuable intellectual property by companies. Model Stealing (MS) attacks allow an adversary with black-box access to a ML model to replicate its functionality by training a clone model using the predictions of the target model for different inputs. However, best available existing MS attacks  fail to produce a high-accuracy clone without access to the target dataset or a representative dataset necessary to query the target model.  In this paper, we show that preventing access to the target dataset is not an adequate defense to protect a model.  
We propose \emph{MAZE} -- a data-free \underline{m}odel stealing \underline{a}ttack using \underline{z}eroth-order gradient \underline{e}stimation that produces high-accuracy clones. In contrast to prior works, MAZE uses only synthetic data created using a generative model to perform MS.


Our evaluation with four image classification models shows that MAZE provides a normalized clone accuracy in the range of $0.90\times$ to $0.99\times$\footnote{Code: \url{https://github.com/sanjaykariyappa/MAZE}}, and outperforms even the recent attacks that rely on partial data (JBDA, clone accuracy $0.13\times$ to $0.69\times$) and on surrogate data (KnockoffNets, clone accuracy $0.52\times$ to $0.97\times$).  We also study an extension of MAZE in the partial-data setting, and develop \emph {MAZE-PD}, which generates synthetic data closer to the target distribution. MAZE-PD further improves the clone accuracy ($0.97\times$ to $1.0\times$) and reduces the query budget required for the attack by $2\times$-$24\times$.

\end{abstract}

\section{Introduction}

\ignore{Advances in Machine Learning (ML) algorithms such as Deep Neural Networks (DNNs) have enabled the automation of a wide variety of challenging tasks in multiple fields, including computer vision and speech recognition. This, in turn, has enabled companies to train and deploy DNN models to create new products and services like intelligent cameras, voice assistants, self-driving cars, and even improve existing products such as web search and predictive text. Furthermore, several companies currently also offer ML as a service where a user can query a remotely hosted ML model with an input to obtain the output predictions. The users of such services typically only have black-box access to the predictions of the model without knowing the model parameters or architecture.}

The ability of Deep Neural Networks (DNNs) to achieve state of the art performances in a wide variety of challenging computer-vision tasks has spurred the wide-spread adoption of these models by companies to enable various products and services such as self-driving cars, license plate reading, disease diagnosis from medical images, activity classification from images and video, and smart cameras. As the performance of ML models scales with the training data~\cite{data}, companies invest significantly in collecting vast amounts of data to train high-performance ML models. Protecting the confidentiality of these  models is vital for companies to maintain a competitive advantage and to prevent the stolen model from being misused by an adversary to compromise security and privacy. For example, an adversary can use the stolen model to craft adversarial examples~\cite{exp_harness, intrigue, transfer}, compromise user membership privacy through membership inference attacks~\cite{membership_inference,yeom2018privacy,nasr2018comprehensive}, and leak sensitive user data used to train the model through model inversion attacks~\cite{model_inversion,dreaming,zhang2019secret}. Thus, ML models are considered valuable intellectual properties of the owner and are closely guarded against theft and data leaks.

\begin{figure}[htb]
 \vspace{-0.05 in}
	\centering
    \centerline{\epsfig{file=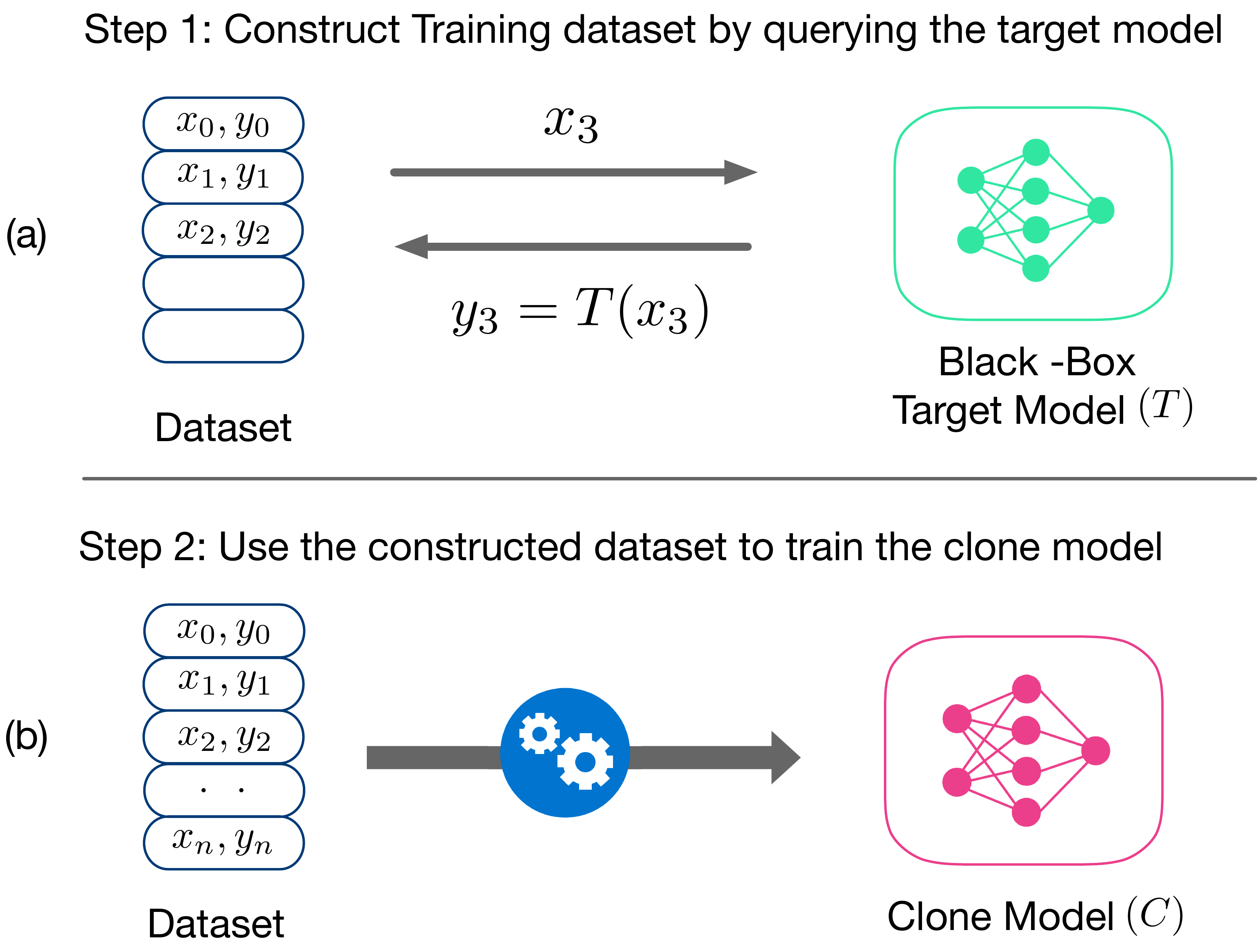, width=0.85\columnwidth}}
    \vspace{-0.05 in}
	\caption{Model stealing attacks: The target model is queried using a set of inputs $\{x_i\}_{i=1}^n$ to obtain a labeled training dataset $\{x_i,y_i\}_{i=1}^n$, which is used to train the clone model. }
    \label{fig:kd}
    \vspace{-0.05 in}
\end{figure}

Model functionality stealing attacks compromise the confidentiality of ML models by allowing an adversary to train a \emph{clone model} that closely mimics the predictions of the target model, effectively copying its functionality. These attacks only require black-box access to the target model where the adversary can access the predictions of the model for any given input. Fig.~\ref{fig:kd} illustrates the steps involved in carrying out a MS attack.  The adversary first queries the target model $T$ with various inputs $\{x_i\}_{i=1}^n$ and uses the predictions of the target model $y_i = T(x_i)$ to construct a labeled dataset $\mathcal{D}=\{x_i, y_i\}$. This dataset is then used to train a clone model $C$ to match the predictions of $T$.

In the current state of the art methods (e.g., \cite{practicalbb,knockoff}), the availability of in-distribution or similar surrogate data to query the target model plays a key role in the ability of the attacker to train high accuracy clone models. However, in most real-world scenarios, the training data is not readily available to the attacker as companies typically train their models using proprietary datasets. To carry out MS in such a data-limited setting, existing attacks either assume partial availability of the target dataset or the availability of a \emph{surrogate} dataset that is semantically similar to the target dataset (e.g., using CIFAR-100 to attack a CIFAR-10 model). For example, {\em Jacobian-Based Dataset Augmentation (JBDA)}~\cite{practicalbb} is an attack that uses a subset of the training data to create additional synthetic data, which is used to query the target model. {\em KnockoffNets}~\cite{knockoff} is another MS attack that uses a \emph{surrogate} dataset to query the target model.  
These attacks become ineffective without access to the target dataset or a representative surrogate dataset. \footnote{We refer the interested readers to Section 6.1 of the KnockoffNets paper~\cite{knockoff} for a discussion on the importance of using semantically similar datasets to carry out the attack.}

This paper is the first to show that a highly accurate MS attack is feasible without relying on any access to the target dataset or even a surrogate dataset -- our method only relies on synthetically-generated out-of-distribution data -- but results in high-accuracy clones on in-distribution data. We make the following key contributions in our paper:
\ignore{This may seem somewhat counter-intuitive, but there is recent work in other contexts that relies on the same principle. For instance, recent works in data-free knowledge distillation~\cite{abm, dream_distillation} have shown that synthetic data can be used to create a student model (typically, a compressed or simplified model for use on low-end computing devices) from a white-box teacher model. }



\textbf{Contribution 1:} We propose \emph{MAZE}-- the first data-free model stealing attack capable of training high-accuracy clone models across multiple image classification datasets and complex DNN target models. In contrast to existing attacks that require some form of data to query the target, MAZE uses synthetic data created using a generative model to carry out MS attack. Our evaluations across DNNs trained on various image classification tasks show that MAZE provides a normalized clone accuracy of $0.90\times$ to $0.99\times$ (normalized clone accuracy is the accuracy of the clone model expressed as a fraction of the target-model accuracy). Despite not using any data, MAZE outperforms recent attacks that rely on partial data (JBDA, clone accuracy of $0.13\times$ to $0.69\times$) or surrogate data (KnockoffNets, clone accuracy of $0.52\times$ to $0.97\times$).

\textbf{Contribution 2:} Our key insight is to draw inspiration from data-free knowledge distillation (KD) and zeroth-order gradient estimation to train the generative model used to produce synthetic data in MAZE. Similar to data-free KD, the generator is trained on a disagreement objective, which encourages it to produce synthetic inputs that maximize the disagreement between the predictions of the target (teacher) and the clone (student) models. 
By training the clone model on such synthetic examples we can improve the alignment of the clone model's decision boundary with that of the target, resulting in a high-accuracy clone model.

In data-free KD, training the generator on the disagreement objective is possible since white-box access to the teacher model is available.  But, unlike in data-free KD, MAZE operates in a black-box setting.  We therefore leverage {\em zeroth-order gradient estimation (ZO)}~\cite{nesterov2017random,ghadimi2013stochastic} to approximate the gradient of the black-box target model and use this to train the generator. Unfortunately, we found a direct application of ZO gradient estimation to be impractical on real-world image classification models since the dimensionality of the generator's parameters can be in the order of millions. We propose a way to overcome the dimensionality problem by estimating gradients with respect to the significantly lower-dimensional synthetic input and show that our method can be successfully used to train a generator in a query-efficient manner.

\textbf{Contribution 3:} In some cases, partial datasets may be available. Recognizing that, we propose an extension of MAZE, called {\em MAZE-PD}, for scenarios where a small partial dataset (e.g., 100 examples) is available to the attacker.  MAZE-PD leverages the available data to produce queries that are closer to the training distribution than in MAZE by using generative adversarial training. Our evaluations show that MAZE-PD provides near-perfect clone accuracy ($0.97\times$ to $1.0\times$), while reducing the number of queries by $2\times$-$24\times$ compared to MAZE. 

In summary, our key finding is that an attacker only requires black-box access to the target model and no in-distribution data to create high-accuracy clone models in the image classification domain. If even a very limited amount of in-distribution data is available, near-perfect clone accuracy is feasible. This raises questions on how machine learning models can be better protected from competitors and bad actors in this domain.

\section{Related Work}
Several types of MS attacks have been proposed in recent literature. Depending on the goal of the attack, MS attacks can be categorized into: (1) parameter stealing (2) hyper-parameter stealing (3) functionality stealing attacks. Parameter stealing attacks~\cite{tramer2016stealing, lowd2005adversarial} focus on stealing the exact model parameters, while hyper-parameter stealing attacks~\cite{wang2018stealing, oh2019towards} aim to determine the hyper-parameters used in the model architecture or the training algorithm of the target model. Our work, MAZE and MAZE-PD,  are designed to carry out a {\em functionality stealing} attack, where the goal is to replicate the functionality of a blackbox target model by training the clone model on the predictions of the target. As the attacker typically does not have access to the dataset used to train the target model, attacks need alternate forms of data to query the target model and perform model stealing. Depending on the availability of data, functionality stealing attacks can be classified as using (1) partial-data, (2) surrogate-data, or (3) data-free, i.e., synthetic data.  We discuss prior works in each of these three settings and also briefly discuss relationship between model stealing and knowledge distillation.

\subsection{Model Stealing with Partial Data}
\label{sec:pd_setting}
In the partial-data setting, the attacker has access to a subset of the data used to train the target model. While this in itself may be insufficient to carry out model stealing, it allows the attacker to craft synthetic examples using the available data. {\em Jacobian Based Dataset Augmentation (JBDA)}~\cite{practicalbb} is an example of one such attack that assumes that the adversary has access to a small set of \emph{seed} examples from the target data distribution. The attack works by first training a clone model $C$ using the seed examples and then progressively adding synthetic examples to the training dataset. JBDA uses a perturbation based heuristic to generate new synthetic inputs from existing labeled inputs. E.g., from an input-label pair $(x, y)$, a synthetic input $x'$ is generated by using the jacobian of the clone model's loss function $\nabla_x\mathcal{L}\left(C\left(x;\theta_c\right),y\right)$ as shown in Eqn.~\ref{eq:jbda}. 

\begin{align} \label{eq:jbda}
x' = x + \lambda sign\left(\nabla_x\mathcal{L}\left(C\left(x;\theta_c\right),y\right)\right)
\end{align}

\vspace{0.1 in}

The dataset of synthetic examples $\{x'_i\}$ generated this way are labeled by using the predictions of the target model $y'_i = T(x'_i)$ and the labeled examples $\{x'_i,y'_i\}$ are added to the pool of labeled examples that can be used to train the clone model $C$. In addition to requiring a set of seed examples from the target distribution, a key limitation of JBDA is that, while it works well for simpler datasets like MNIST, it tends to produce clone models with lower classification accuracy for more complex datasets. For example, our evaluations in Section~\ref{sec:experiments} show that JBDA provides a normalized clone accuracy of only $0.13\times$ (GTSRB dataset) and $0.18\times$ (SVHN dataset).

\subsection{Model Stealing with Surrogate Data}
In the surrogate data setting, the attacker has access to alternate datasets that can be used to query the target model. \ignore{For example, consider an attacker who wants to steal a DNN model trained with the CIFAR-10 dataset. Given a lack of access to a large collection of examples from CIFAR-10, the attacker can potentially use images from an alternate dataset, such as CIFAR-100 to query the target model to perform model stealing.} KnockoffNets~\cite{knockoff} is an example of a MS attack that is designed to operate in such a setting. With a suitable surrogate dataset, KnockoffNets can produce clone models with up to $0.97\times$ the accuracy of the target model. However, the efficacy of such attacks is dictated by the availability of a suitable surrogate dataset. For instance, if we use the MNIST dataset to perform MS on a FashionMNIST model, it only produces a clone model with $0.41\times$ the accuracy of the target model (See Table~\ref{table:results} for full results).
This is because the surrogate dataset is not representative of the target dataset, which reduces the effectiveness of the attack.

\subsection{Data-Free Model Stealing}

In the data-free setting, the adversary does not have access to any data. This represents the hardest setting to carry out MS as the attacker has no knowledge of the data distribution used to train the target model. A recent work by Roberts et al.~\cite{msnoise} studies the use of inputs derived from various noise distributions to carry out MS attack in the data-free setting. While this attack works well for simple datasets like MNIST, our evaluations show that such attacks do not scale to more complex datasets such as CIFAR-10 (we obtained relative clone accuracy of only $0.11\times$), limiting their applicability (See Table~\ref{table:results} for full results). 

\subsection{Knowledge distillation}
Model stealing is related to  knowledge-distillation (KD)~\cite{hinton2015distilling}, but in KD, unlike in model stealing, the target model is available to the attacker and is simply being summarized into a simpler architecture. Appendix~\ref{app:related_dfkd} further discusses works in data-free KD and explain why the these works are not directly applicable for MS attacks.

\begin{figure*}[tb]
	\centering
    \centerline{\epsfig{file=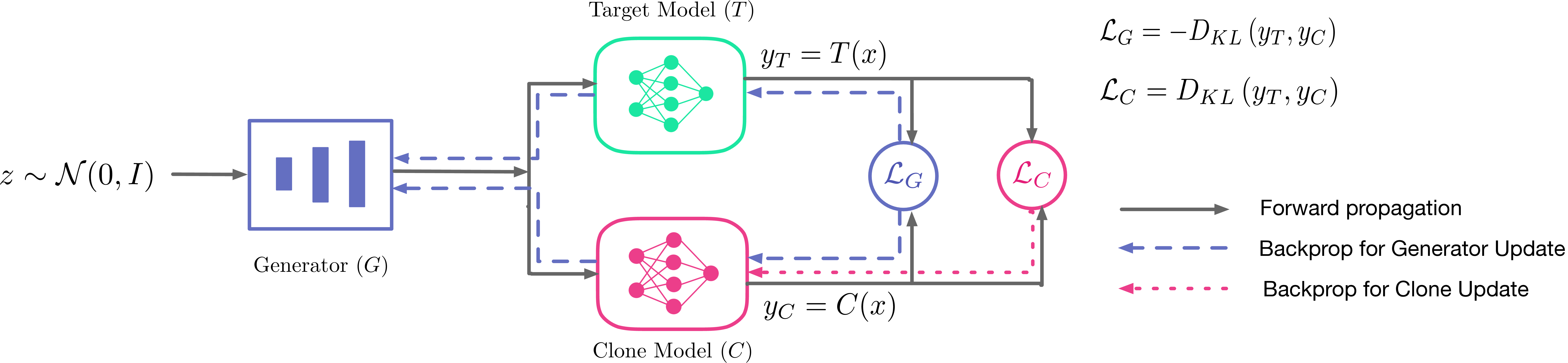, width=0.94\textwidth}}
    \caption{MAZE Attack Setup: MAZE uses a generative model $G$ to produce the synthetic input queries $\{x\}$ to perform Model Stealing. The clone model $C$ is trained to match the predictions of the target model $T$. $G$ is trained to produce queries that maximize the dissimilarity between $y_T$ and $y_C$. Optimizing $\mathcal{L}_G$ requires backpropagation through $T$ to update $G$. However, we only have black-box access to $T$, therefore we use zeroth-order gradient estimation to perform gradient descent on $\mathcal{L}_G$.}
    \vspace{-0.1 in}
    \label{fig:zoom}
\end{figure*}

\section{Preliminaries}
The goal of this paper is to develop a model functionality stealing attack in the data-free setting, which can be used to train a high-accuracy clone model only using black-box access to the target model. We formally state the objective and constraints of our proposed model stealing attack.

\vspace{0.1 in}

\textbf{Attack Objective:}
Consider a target model $T$ that performs a classification task with high accuracy. Our goal is to train a clone model $C$ that replicates the functionality of the target model by maximizing the accuracy on a test set $\mathcal{D}_{test}$ as shown in Eqn. ~\ref{eq:adv_obj}.

\begin{align} \label{eq:adv_obj}
\max_{\theta_C}\mathop{\mathbb{E}}_{x, y\sim \mathcal{D}_{test}}[Acc(C(x;\theta_C), y)]
\end{align}

\vspace{0.1 in}

\textbf{Attack Constraints:}
We assume that the adversary does not know any details about the Target model's architecture or the model parameters $\theta_T$. The adversary is only allowed black-box access to the target model. We assume the \emph{soft-label} setting where the adversary can query the target model with any input $x$ and observe its output probabilities $\vec{y}=T(x;\theta_T)$. We consider model stealing attacks under two settings based on the availability of data: 

\vspace{0.05 in}

1. \emph{Data-free setting (Primary goal):} The adversary does not have access to the dataset $\mathcal{D}_{T}$  used to train the target model or a good way to sample from the target data distribution $\mathbb{P}_{T}$. (Section~\ref{sec:maze})

2. \emph{Partial-data setting (Secondary goal):} The adversary has access to a small subset (e.g., 100) of training examples randomly sampled from the training dataset of the target model. (Section~\ref{sec:maze-pd})

For both of these settings we assume the availability of a test set $\mathcal{D}_{test}$, which is used to report the test accuracies of the clone models produced by our attack.

\section{MAZE: Data-Free Model Stealing} ~\label{sec:maze}
We propose \emph{MAZE}, a data-free \underline{m}odel stealing \underline{a}ttack using \underline{z}eroth order gradient \underline{e}stimation. Unlike existing attacks, MAZE does not require access to the target or a surrogate dataset and instead uses a generative model to produce the synthetic queries for launching the attack. Fig.~\ref{fig:zoom} shows an overview of MAZE.\ignore{Similar to prior works on Model Stealing, } \ignore{MAZE trains a clone model $C$ using the predictions of $T$ for various input queries $\{x\}$. MAZE uses a generative model $G$ to produce the inputs necessary to query the target model.} In this section, we first describe the training objectives of the clone and the generator model. We then motivate the need for gradient estimation to update $G$ in the black-box setting of MS attack and show how zeroth-order gradient estimation can be used to optimize the parameters of $G$. Finally, we discuss our algorithm to carry out model stealing with MAZE.

\subsection{Training the Clone Model}
 The clone model is trained using the input queries produced by the generator. The generator $G$ takes in a low dimensional latent vector $z$, sampled from a random normal distribution, and produces an input query $x\in\mathbb{R}^d$ that matches the input dimension of the target classifier (Eqn.~\ref{eq:fwprop_gen}). We use $x$ to obtain the output probabilities of the target model $\vec{y_T}$ and clone model $\vec{y_C}$ on $x$ as shown in Eqn.~\ref{eq:fwprop2}.
 \vspace{-0.01 in}
\begin{align}
x &= G(z; \theta_G);\ \ z \sim \mathcal{N}(0, \boldsymbol{I})\label{eq:fwprop_gen}\\
\vec{y_T} &= T(x; \theta_T);\ \vec{y_C} = C(x; \theta_C)\label{eq:fwprop2}
\end{align}

Where $\theta_T$, $\theta_C$ and $\theta_G$ represent the parameters of the target, clone, and generator models, respectively. The clone model is trained using the loss function in Eqn.~\ref{eq:loss_c} to minimize the KL divergence between $\vec{y_C}$ and $\vec{y_T}$. 
 \vspace{-0.05 in}
\begin{align}
\mathcal{L}_C = D_{KL}(\vec{y_T} \| \vec{y_C})\label{eq:loss_c}    
\end{align}

\subsection{Training the Generator Model}
The generator model $G$ synthesises the queries necessary to perform model stealing. Similar to recent works in data-free KD~\cite{abm,dreaming, data_free_adv_dist}, MAZE trains the generator to produce queries that maximize the disagreement between the predictions of the teacher and the student by maximizing the KL-divergence between $\vec{y_T}$ and $\vec{y_C}$. The loss function used to train the generator model is described by Eqn.~\ref{eq:loss_g}, which we refer to as the \emph{disagreement objective}.  
\begin{align}
\mathcal{L}_G = -D_{KL}(\vec{y_T} \| \vec{y_C})\label{eq:loss_g}    
\end{align}
Training $G$ on this loss function maximizes the disagreement between the predictions of the target and the clone model. Since $C$ and $G$ have opposing objectives, training both models together results in a two-player game, similar to \emph{Generative Adversarial Networks}~\cite{gan}, resulting in the generation of inputs that maximize the learning of the clone model. By training $C$ to match the predictions of $T$ on the queries generated by $G$, we can  perform knowledge distillation and obtain a highly accurate clone model.

Training $G$ using the loss function in Eqn.~\ref{eq:loss_g} requires backpropagating through the predictions of the target model $T$, as shown by the dashed lines in Fig.~\ref{fig:zoom}. Unfortunately, as we only have black-box access to $T$, we cannot perform back-propagation directly, preventing us from training G and carrying out the attack. To solve this problem, our insight is to use {\em zeroth-order gradient estimation} to approximate the gradient of the loss function $\mathcal{L}_G$. The number of black-box queries necessary for ZO gradient estimation scales with the dimensionality of the parameters being optimized. 
Estimating the gradients of $\mathcal{L}_G$ with respect to the generator parameters $\theta_G$ directly is expensive as the generator has on the order of millions of parameters. Instead, we choose to estimate the gradients with respect to the synthetic input $x$ produced by the generator, which has a much lower dimensionality ($3072$ for CIFAR-10), and use this estimate to back propagate through $G$. This modification allows us to compute a gradient estimates in a query efficient manner to update the generator model. The following section describes how we efficiently apply zeroth-order gradient estimation to train the generator model.

\subsection{Train via Zeroth-Order Gradient Estimate} \label{sec:gen_training}
Zeroth-order gradient estimation~\cite{nesterov2017random,ghadimi2013stochastic} is a popular technique to perform optimization in the black-box setting. We use this technique to train our generator model $G$. Recall that our objective is to update the generator model parameters $\theta_G$ using gradient descent to minimize the loss function $\mathcal{L}_G$ as shown in Eqn.~\ref{eq:update_g}.
\begin{align}
\theta_G^{t+1} = \theta_G^{t} - \eta \nabla_{\theta_G} \mathcal{L}_G \label{eq:update_g}
\end{align}

Updating $\theta_G$ in this way requires us to compute the derivative of the loss function $\nabla_{\theta_G}\mathcal{L}_G$. By the use of chain-rule, $\nabla_{\theta_G}\mathcal{L}_G$ can be decomposed into two components as shown in Eqn.~\ref{eq:loss_g_chain}.

\begin{align}
\nabla_{\theta_G}\mathcal{L}_G = \frac{\partial \mathcal{L}_G}{\partial \theta_G} = \frac{\partial \mathcal{L}_G}{\partial x} \times \frac{\partial x}{\partial \theta_G}\label{eq:loss_g_chain}    
\end{align}

We can compute the second term $\frac{\partial x}{\partial \theta_G}$  in Eqn.~\ref{eq:loss_g_chain} by performing backpropagation through $G$. Computing the first term $\frac{\partial \mathcal{L}_G}{\partial x}$ however requires access to the model parameters of the target model ($\theta_T$). Since $T$ is a black-box model from the perspective of the attacker, we do not have access to $\theta_T$, which prevents us from computing $\frac{\partial \mathcal{L}_G}{\partial x}$ through backpropagation. Instead, we propose to use an approximation of the gradient by leveraging zeroth-order gradient estimation. To explain how the gradient estimate is computed, consider an input vector $x \in \mathbb{R}^{d}$ generated by $G$ that is used to query $T$. We can estimate $\frac{\partial \mathcal{L}_G}{\partial x}$ by using the method of forward differences~\cite{fw_diff} as shown in Eqn.~\ref{eq:fw_diff1}. 

\begin{align}
\hat{\nabla}_x \mathcal{L}_G(x;u_i) = \frac{d\cdot\left(\mathcal{L}_{G}\left(x+\epsilon u_i\right)-\mathcal{L}_{G}(x)\right)}{\epsilon} u_i \label{eq:fw_diff1}    
\end{align}

Where $u_i$ is a random variable drawn from a $d$ dimensional unit sphere with uniform probability and $\epsilon$ is a small positive constant called the {\em smoothing factor}. 
The random gradient estimate, shown in Eqn.~\ref{eq:fw_diff1}, tends to have a high variance. To reduce the variance, we use an averaged version of the random gradient estimate ~\cite{biased_estimate1,variance1} by computing the forward difference using $m$ random directions $\{u_1,u_2,..u_m\}$,  as shown in Eqn.~\ref{eq:fw_diff2}. 
\begin{align}
\hat{\nabla}_x \mathcal{L}_G(x) = \frac{1}{m} \sum_{i=1}^{m} \hat{\nabla}_x \mathcal{L}_G(x;u_i)  \label{eq:fw_diff2}    
\end{align}

Where $\hat{\nabla}_x \mathcal{L}_G$ is an estimate of the true gradient $\nabla_x \mathcal{L}_G$. By substituting $\hat{\nabla}_x \mathcal{L}_G$ into Eqn.~\ref{eq:loss_g_chain}, we can compute an approximation for the gradient of the loss function of the generator: $\hat{\nabla}_{\theta_G} \mathcal{L}_G$. The gradient estimate $\hat{\nabla}_{\theta_G} \mathcal{L}_G$ computed this way can be used to perform gradient descent by updating the parameters of the generator model $\theta_G$ according to Eqn.~\ref{eq:update_g}. By updating $\theta_G$, we can train $G$ to produce the synthetic examples required to perform model stealing. 

\ignore{One problem that arises with numerical approximation of gradients is that we need to ensure that the inputs used to query the target model should have values that lie within the valid range, in our case $\left[-1,1\right]$. However, the perturbation $(x+\epsilon u_i)$ might cause the input to take values outside this range, making the query invalid. To avoid this problem, we apply the gradient estimation to the activations $x_p$ in the penultimate layer of $G$, which does not have a constraint on the values it can take. Note that the last layer of $G$ consists of a $tanh$ function that ensures that the output of the generator $x=tanh(x_p)$ is always in the range $\left[-1,1\right]$.}

\subsection{MAZE Algorithm for Model Stealing Attack}

We outline the algorithm of MAZE in Algorithm~\ref{alg:attack} by putting together the individual training algorithms of the generator and clone models. We start by fixing a query budget $Q$, which dictates the maximum number of queries we are allowed to make to the target model $T$. $\epsilon$ is the smoothing parameter and $m$ is the number of random directions used to estimate the gradient. We set the value of $\epsilon$ to $0.001$ in our experiments. 
$N_G,N_C$ represent the number of training iterations and $\eta_G,\eta_C$ represent the learning rates of the generator and clone model, respectively. $N_R$ denotes the number of iterations for experience replay. 

\ignore{Our attack starts by initializing the generator and the clone model $G(\cdot;\theta_G), C(\cdot;\theta_C)$. $q$ is a variable that holds the number of queries already performed and $\mathcal{D}$ is a dataset used to collect the input, label pairs: $\left(x,T(x)\right)$ generated by querying the target model.}

\begin{algorithm} 
\SetAlgoLined 
\KwIn{$T,Q,\epsilon,m,N_G,N_C,N_G,\eta_G,\eta_C$}
\KwOut{Clone model $C(\cdot;\theta_C)$}
 Initialize $G(\cdot;\theta_G), C(\cdot;\theta_C), q\gets0, \mathcal{D}\gets\{\}$

 \While{$q<Q$}{

 // Generator Training
 
 \For{$i\gets0$ \KwTo $N_G$}{
    $x=G(z): z\sim\mathcal{N}(0,I)$
    
    $\mathcal{L}_G = -D_{KL}\left(T(x)\|C(x)\right)$
    
    $\hat{\nabla}_{\theta_G}\mathcal{L}_G \gets ZO\_grad\_est(G,T,C,x,\epsilon,m)$
    
    
    $\theta_G \gets \theta_G - \eta_G \hat{\nabla}_{\theta_G}\mathcal{L}_G$

  
  }
  
  // Clone Training
  
  \For{$i\gets0$ \KwTo $N_C$}{
    $x=G(z): z\sim\mathcal{N}(0,I)$
    
    $\mathcal{L}_C = D_{KL}\left(T(x)\|C(x)\right)$
    
    $\theta_C \gets \theta_C - \eta_C \nabla_{\theta_C}\mathcal{L}_C$

    $\mathcal{D} \gets \mathcal{D}\cup\{(x,T(x))\}$
  
  }
  
  // Experience Replay

  \For{$i\gets0$ \KwTo $N_T$}{
  
    $(x,y_T) \sim \mathcal{D}$
    
    $\mathcal{L}_C = D_{KL}\left(y_T\|C(x)\right)$
    
    $\theta_C \gets \theta_C - \eta_C \nabla_{\theta_C}\mathcal{L}_C$
  
  }
  $q\gets update(q)$
 }

 \caption{MAZE Algorithm for Model Stealing Attack}
 \label{alg:attack}
\end{algorithm}

The outermost loop of the attack repeats till we exhaust our query budget $Q$. The attack algorithm involves three phases: 1. \emph{Generator Training} 2. \emph{Clone Training} and 3. \emph{Experience Replay}. In the \emph{Generator Training} phase, we perform $N_G$ rounds of gradient descent for $G$, which is trained to produce inputs $x$ that maximize the KL-divergence between the predictions of the target and clone model. $\theta_G$ is updated by using zeroth-order gradient estimates as described in Section~\ref{sec:gen_training}. This is followed by the \emph{Clone Training} phase, where we perform $N_C$ rounds of gradient descent for $C$. In each round, we generate a batch of inputs $x=G(z)$ and use these inputs to query the target model. The clone model is trained to match the predictions of the target model by minimizing $D_{KL}(T(x)\|C(x))$. The input, prediction pair: $(x,T(x))$ generated in each round is stored in dataset $\mathcal{D}$. Finally, we perform \emph{Experience Replay}, where we train the clone on previously seen inputs that are stored in $\mathcal{D}$. Retraining on previously seen queries reduces \emph{catastrophic forgetting}~\cite{catastrophic} and ensures that the clone model continues to classify old examples seen during the earlier part of the training process correctly.

\subsection{Computing the Query Cost} 

The target model needs to be queried in order to update both the generator and the clone models. Considering a batch size of 1, one training iteration of $G$ requires $m +1$ queries to $T$ for the zeroth-order gradient estimation and each training loop of $C$ requires $1$ query. Experience replay, on the other hand, does not require any additional queries to $T$. Thus, with a batch size of $B$, the query cost of each iteration is described by Eqn.~\ref{eq:query_cost}
\begin{align}
\text{Query cost per iteration} = B (N_G(m+1) + N_C)
  \label{eq:query_cost}    
\end{align}

We use $B=128$, $N_G=1, N_C=5, N_R=10$ and $m=10$ in our experiments, unless stated otherwise. Thus, each iteration of the attack requires $2048$ queries. We use a query budget of $5M$ for FashionMNIST and SVHN and a query budget of $30M$ for GTSRB and CIFAR-10 datasets to report our results. 
\section{Experimental Evaluation}\label{sec:experiments}

We validate our attack by performing model stealing on various target models and provide experimental evidence to show that our attack can produce high accuracy clone models without using any data. 
We compare our results against two prior works-- \emph{KnockoffNets} and \emph{Jacobian Based Dataset Augmentation} (JBDA)-- and show that the clone models produced by our attack have comparable or better accuracy than the ones produced by these prior works, despite not using any data. In addition, we also perform sensitivity studies to understand the impact of various attack parameters including query cost and number of gradient estimation directions in Appendix~\ref{app:sensitivity}.   



\subsection{Setup: Dataset and Architecture}

We perform our evaluations by attacking DNN models that are trained on various image-classification tasks. The datasets and target model accuracies used in our experiments are mentioned in Table~\ref{table:results}. We use a LeNet for the FashionMNIST and ResNet-20 for the other datasets as the target model. Our attack assumes no knowledge of the target model and uses a randomly initialized 22-layer WideResNet~\cite{wresnet} as the clone model for all the datasets. In general, any sufficiently complex DNN can be used as the clone model. We use an SGD optimizer with an initial learning rate of $0.1$ to train our clone model. 
For $G$, we use a generative model with 3 convolutional layers. Each convolutional layer in $G$ is followed by a batchnorm layer and the activations are upsampled to ensure that the outputs generated by $G$ are of the correct dimensionality corresponding to the dataset being attacked. We use an SGD optimizer with an initial learning rate of $0.0001$  to train $G$. The learning rates for both the clone and generator models are decayed using cosine annealing. 

\ignore{

\begin{table}[htb]
\caption{Dataset, model architecture, and accuracy of the target models $T$ used in our experiments. }
\vspace{-0.1 in}
\begin{center}
\begin{tabular}{|c||c|c|}
\hline
Dataset           & Target DNN Arch. ($T$)     &  Accuracy(\%) \\
\hline\hline
FashionMNIST      & LeNet               & $91.04$        \\
SVHN              & ResNet-20           & $95.25$        \\
GTSRB            & ResNet-20           & $97.43$        \\
CIFAR-10          & ResNet-20           & $92.26$        \\
\hline
\end{tabular}
\end{center}
\label{table:datasets}
\vspace{-0.1in}
\end{table}

}

\begin{table*}[tb]
  \centering
   \caption{Comparison of clone accuracies obtained from various attacks. Numbers in the bracket express the accuracy as a multiple of the target model accuracy. MAZE obtains high accuracy ($0.90 \times$ to $0.99\times$), despite not using any data.}
  \label{table:results}
  \begin{tabular}{|c|c||c|c|c|c|c|c|}
    \hline 
    \multirow{2}{*}{\textbf{Dataset}} & \textbf{Target} & \textbf{MAZE} & \textbf{KnockoffNets} & \textbf{JBDA}& \textbf{Noise}\\
    &\textbf{Accuracy (\%)} &\textbf{(data-free)}&\textbf{(surrogate data)}&\textbf{(partial-data)}& \textbf{(data-free)}\\
    \hline \hline
    FashionMNIST & $91.04$ & $\boldsymbol{81.9\ (0.90\times)}$ & $47.26\ (0.52\times)$ & $62.65\ (0.69\times)$ & $62.91\ (0.69\times)$ \\ \hline
    SVHN & $95.25$ & $\boldsymbol{93.85 (0.99 \times)}$ & $92.77\ (0.97\times)$ & $17.16(0.18\times)$ & $51.86\ (0.54\times)$ \\ \hline
    GTSRB & $97.43$ & $88.31\ (0.91\times)$ & $\boldsymbol{89.86\ (0.92\times)}$ & $12.80(0.13\times)$ & $38.38\ (0.39\times)$ \\ \hline
    CIFAR-10 & $92.26$ & $\boldsymbol{89.85\ (0.97\times)}$ & $82.56\ (0.89\times)$ & $25.11\ (0.27\times)$ & $10.17\ (0.11\times)$ \\ \hline
  \end{tabular}
 
\end{table*}

\subsection{Configuration of Existing Attacks}

Existing MS attacks either use surrogate data or synthetic datasets derived from partial access to the target dataset. We compare MAZE with the following attacks:

\emph{1. KnockoffNets~\cite{knockoff}} attack uses a surrogate dataset to query the target model to construct a labeled dataset using the predictions of the target model. This labeled dataset is used to train the clone model. We use MNIST, CIFAR10, CIFAR100, and CIFAR10 as the surrogate datasets for FashionMNIST, SVHN, CIFAR10, and GTSRB models, respectively. In each case, we query the target model with the training examples of the surrogate dataset. We then use the dataset constructed from these queries to train the clone model for 100 epochs using an SGD optimizer with a learning rate of $0.1$ with cosine annealing scheduler.

\emph{2. JBDA~\cite{practicalbb}:} attack performs MS by using synthetic examples to query the target model. These synthetic examples are generated by adding perturbations to a set of \emph{seed} examples, which are obtained from the data distribution of the target model. The perturbations are computed using the Jacobian of the clone model's loss function (Eqn.~\ref{eq:jbda}).  We start with an initial dataset of 100 seed examples and perform 6 rounds\footnote{We found that the accuracy of the JBDA attack stagnates beyond 6 augmentation rounds. This is in line with the observations made by Juuti et al.~\cite{prada}.} of synthetic data augmentation with the clone model being trained for 10 epochs between each round. $\lambda$ in Eqn.~\ref{eq:jbda} dictates the magnitude of the perturbation. We set this to a value of $0.1$. We use Adam optimizer with a learning rate of 0.001 to train the clone model.
\emph{3. Noise: } To test if inputs sampled from noise can be used to carry out MS attack, we design a \emph{Noise} attack. We follow the proposal by Roberts et al.~\cite{msnoise} and use random samples from an Ising prior model to query the target model. This attack serves as a baseline data-free MS attack to compare with our proposal.

\subsection{Key Result: Normalized Clone Accuracy}
Table~\ref{table:results} shows the clone-accuracy obtained by attacking various target models using MAZE. The numbers in brackets express the clone accuracy normalized to the accuracy of the target model being attacked. We also compare MAZE with existing MS attacks and highlight the best clone accuracy for each dataset in bold.
Our results show that MAZE produces high accuracy clone models with a normalized accuracy greater than $0.90\times$ for all the target models under attack. In contrast, the baseline \emph{Noise} attack fails to produce high accuracy clone models for most of the datasets.

Furthermore, the results from our attack also compare favorably against \emph{KnockoffNets} and \emph{JBDA}, both of which require access to some data. We find that the effectiveness of KnockoffNets is highly dependent on the surrogate data being used to query the target model. For example, using MNIST to attack FashionMNIST dataset results in a low accuracy clone model ($0.52\times$ target accuracy) as these datasets as visually dissimilar. However, using CIFAR-100 to query CIFAR-10 results in a high accuracy clone model ($0.89\times$ target accuracy) due to the similarities in the two datasets.
JBDA seems to be effective for attacking simpler datasets like $FashionMNIST$, but the accuracy reduces when attacking more complex datasets. This is in part because JBDA produces queries that are highly correlated to the initial set of ``seed" examples, which sometimes results in worse performance even compared to noise (e.g. SVHN). By using the disagreement objective to train the generator, MAZE  can generate queries that are more useful in training the clone model and result in higher accuracy of clones ($0.91\times$-$0.99\times$) compared to other attacks like JBDA ($0.13\times$-$0.69\times$) that use synthetic data.

\section{MAZE-PD: MAZE with Partial-Data}~\label{sec:maze-pd}

\ignore{Thus far, we have described our MS attack in a \emph{data-free} setting, where the adversary does not have access to any dataset.} The accuracy and speed of our attack can be improved if a few examples from the training-data distribution of the target model are available to the adversary. In this section we develop {\em MAZE-PD}, an extension of MAZE to the partial-data setting.\ignore{ MAZE-PD uses an organization similar to MAZE, where a generator model is trained to create synthetic data. However, MAZE-PD uses {\em Waserstein Generative Adversarial Networks (WGANs)}~\cite{wgan} to train the generator in order to improve the quality of images generated. } In the data-free setting of MAZE, $G$ is trained on a \emph{disagreement objective} to produce inputs that maximize the disagreement between the target and the clone model. In the presence of a limited amount of data, we can additionally train the generator to produce inputs that are closer to the target distribution by using the {\em Waserstein Generative Adversarial Networks (WGANs)}~\cite{wgan} training objective. We observe that even a small amount of data from the target distribution ($100$ examples) can enable the  generator to produce synthetic inputs that are closer to the target distribution (see Appendix~\ref{app:synthetic_images} for example images). By improving the quality of the generated queries, MAZE-PD not only improves the effectiveness of the attack but also allows the attack to succeed with far fewer queries compared to MAZE. In this section, we describe how WGANs can be incorporated into the training of the generator model to develop MAZE-PD. We also provide empirical evidence to show that MAZE-PD improves clone accuracy and reduces query cost significantly compared to MAZE.

\ignore{we first provide background on WGANs, which can be used to generate synthetic examples closer to the target distribution. Next, we describe how WGANs can be incorporated into the training of the generator model to develop MAZE-PD. Finally, we provide empirical evidence to show that MAZE-PD improves clone accuracy and reduces query cost significantly compared to MAZE.}

\begin{figure*}[tb]
	\centering
	\vspace{0.05 in}
    \centerline{\epsfig{file=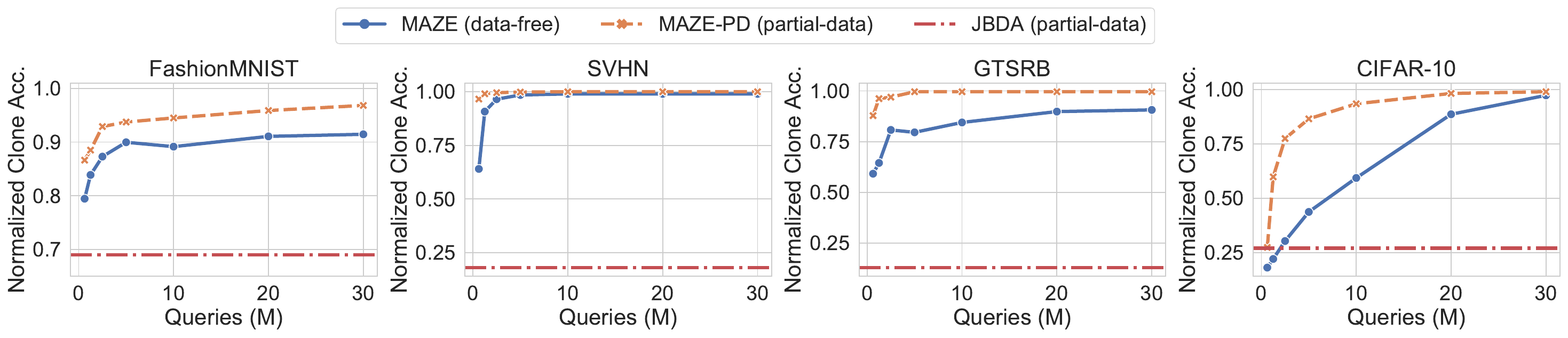, width=\textwidth}}
	\caption{Normalized clone accuracy of MAZE (data-free), MAZE-PD (partial-data), and JBDA (partial-data) as the query budget is varied. Our results show that for a given query budget, MAZE-PD can train a clone model with higher accuracy than MAZE. The accuracy of MAZE-PD is also significantly better than the JBDA attack.}
    \label{fig:query_sensitivity_pd}
\end{figure*}

\subsection{Incorporating WGAN in MAZE}

We describe the modifications to the  training algorithm of MAZE (Algorithm~\ref{alg:attack})  to incorporate WGAN training in the partial-data setting. In addition to the generator ($G$) and clone ($C$) models, we define a \emph{critic model} $D$, which estimates the Wasserstein distance between the target data distribution $\mathbb{P}_{T}$ and the synthetic data distribution of the generator $\mathbb{P}_{G}$ using the function described in Eqn.~\ref{eq:wganx}.

\begin{align}\label{eq:wganx}
W\left(\mathbb{P}_{T}, \mathbb{P}_{G}\right)= \max_{\theta_D} \mathbb{E}_{x \sim \mathbb{P}_T}\left[D(x)\right]-\mathbb{E}_{z \sim \mathcal{N}(0,I)}\left[D\left(G(z)\right)\right]
\end{align}

The generator model aims to produce examples closer to the target distribution by minimizing the Wasserstein distance estimated by the critic model. To incorporate the WGAN objective in the training of the generator, we modify the original loss function of $G$ (Eqn.~\ref{eq:loss_g}) with an additional term as shown in Eqn.~\ref{eq:loss_g_wgan}. 

\begin{align}
x &= G(z); z\sim\mathcal{N}(0,I) \nonumber\\
\mathcal{L}_G &= -D_{KL}(T(x) \| C(x)) -\lambda D(x) \label{eq:loss_g_wgan}
\end{align}

The first term in Eqn.~\ref{eq:loss_g_wgan} represents the disagreement loss from MAZE and the second term is the WGAN loss. The hyper-parameter $\lambda$ balances the relative importance between these two losses. To train the \emph{critic model} $D$, we add an extra training phase (described by Algorithm~\ref{alg:discriminator}) to our original training algorithm. We also include a gradient penalty term $GP=\left(\left\|\nabla_{x} D(x)\right\|_{2}-1\right)^{2}$ in $\mathcal{L}_D$  to ensure that $D$ is  $1$-Lipschitz continuous. We refer the reader to Appendix~\ref{app:wgan} for a more detailed explanation of WGANs.

\begin{algorithm} 
\SetAlgoLined 

  // Critic Training
  
  \For{$i\gets0$ \KwTo $N_d$}{
    $z\sim\mathcal{N}(0,I); x \sim \{x_i\}_{i=1}^n$
    
    $\mathcal{L}_{D}=D(G(z))-D(x) + GP$
    
    $\theta_D \gets \theta_D - \eta_D \nabla_{\theta_D}\mathcal{L}_D$

  }

 \caption{Critic Training}
 \label{alg:discriminator}
\end{algorithm}

The training loops for the \emph{clone training} and \emph{experience replay} in Algorithm~\ref{alg:attack} remain unchanged. Using these modifications we can train a generator model that produces inputs closer to the target distribution $\mathbb{P}_T$. \ignore{These inputs are more effective in carrying out the task of model stealing and help improve the query efficiency of our attack. Thus, the availability of a few samples from the training data allows MAZE-PD to not only obtain higher accuracy but also to orchestrate a faster attack.  We provide experimental results for MAZE-PD in the next subsection.}


\subsection{Results: Clone Accuracy with Partial Data}
We repeat the MS attack using MAZE-PD in the partial data setting. We assume that the attacker now has access to 100 random examples from the training data of the target model, which is roughly $0.2\%$ of the total training data used to train the target model. We use $\lambda=10$ in Eqn.~\ref{eq:loss_g_wgan} and $N_d=10$ in Algorithm ~\ref{alg:discriminator}. Note that critic training does not require extra queries to the target model. The rest of the parameters are kept the same as before. Fig~\ref{fig:query_sensitivity_pd} shows our results comparing the normalized clone accuracy obtained with MAZE-PD and MAZE (data-free) for various query budgets. For a given query budget, MAZE-PD obtains a higher clone accuracy compared to MAZE and achieves near-perfect clone accuracy ($0.97\times$-$1.0\times$) for all the datasets. Additionally, MAZE-PD offers a reduction of $2\times$ to $24\times$ in the query budget compared to MAZE for a given clone accuracy (see Appendix~\ref{app:maze_pd_query_budget}).

\textbf{Comparison with JBDA:} We compare the performance of MAZE-PD with JBDA, which also operates in the partial-data setting. JBDA produces low clone accuracies for most datasets (less than $0.30\times$ for SVHN, GTSRB, and CIFAR-10). In contrast, MAZE-PD obtains  highly accurate clone models ($0.97\times$-$1.0\times$) across all four datasets.


\section{Conclusion}

\ignore{Commercially available machine learning models are often trained with significant resources and proprietary data, and are considered valuable intellectual property. Model Stealing (MS) attacks can allow the users of such proprietary models to copy the functionality of these models to a clone model. The efficacy of current MS attacks is heavily reliant on the availability of data from the target distribution or a representative distribution. }

This paper proposes {\em MAZE}, a high-accuracy MS attack that requires no input data. To the best of our knowledge, MAZE is the first data-free MS attack that works effectively for complex DNN models trained across multiple image-classification tasks. MAZE uses a generator trained with zeroth-order optimization to craft synthetic inputs, which are then used to copy the functionality of the target model to the clone model. Our evaluations show that MAZE produces clone models with high classification accuracy ($0.90\times$ to $0.99\times$). Despite not using any data, MAZE outperforms recent attacks that rely on partial-data or surrogate-data. Our work presents an important step towards developing highly accurate data-free MS attacks.

In addition, we propose \emph{MAZE-PD} to extend MAZE to the partial-data setting, where the adversary has access to a small number of examples from the target distribution. MAZE-PD uses generative adversarial training to produce inputs that are closer to the target distribution. This further improves accuracy ($0.97\times$ to $1.0\times$) and yields a significant reduction in the number of queries ($2\times$ to $24\times$) necessary to carry out the attack compared to MAZE. 

\ignore{We hope that it serves as a baseline for future works to develop more efficient model stealing attacks in the data-free setting. Furthermore, while we develop and evaluate MAZE in the setting of model-stealing attacks, MAZE is a generalized framework, which enables knowledge distillation in the black-box setting, without requiring any input data. }
\section{Acknowledgements}
We thank M. Emre Gursoy, Ryan Feng, Poulami Das and Gururaj Saileshwar for their feedback. This work was partially supported by a gift from Facebook and is partially based upon work supported by the National Science Foundation under grant numbers 1646392 and 2939445 and under DARPA award 885000. We thank NVIDIA for the donation of the Titan V GPU that was used for this research.

{\small
\bibliographystyle{ieee_fullname}
\bibliography{references}
}
\clearpage
\appendix

\section{Background on WGAN}~\label{app:wgan}
WGANs can be used to train a generative model to produce synthetic examples from a target distribution $\mathbb{P}_T$ using a small set of examples $\{x_i\}_{i=1}^{n}$ sampled from this distribution. To explain the training process, we consider a generative model $G$ parameterized by $\theta_G$, which produces samples $x=G(z;\theta_G)$ where $z\sim\mathcal{N}(0,I)$. Let $\mathbb{P}_{G}$ be the probability distribution of the examples $x$ generated by $G$. WGAN aims to minimize the \emph{Wasserstein Distance} between the generator distribution $\mathbb{P}_{G}$ and the target distribution $\mathbb{P}_T$ by optimizing over the generator parameter $\theta_G$. The expression for the Wasserstein Distance between $\mathbb{P}_{T}$ and $\mathbb{P}_{G}$ is given by Eqn ~\ref{eq:wgan1}.

\begin{align}\label{eq:wgan1}
W\left(\mathbb{P}_{t}, \mathbb{P}_{g}\right)=\inf _{\gamma \in \Pi\left(\mathbb{P}_{t}, \mathbb{P}_{g}\right)} \mathbb{E}_{(x, y) \sim \gamma}[\|x-y\|]
\end{align}

Here, $\Pi\left(\mathbb{P}_{t}, \mathbb{P}_{g}\right)$ denotes the set of all joint distributions $\gamma$ for which $\mathbb{P}_{t}$ and $\mathbb{P}_{g}$ are marginals. Unfortunately, computing the infimum in Eqn ~\ref{eq:wgan1} is intractable as it involves searching through the space of all possible joint distributions $\gamma$. Instead, Arjovsky et al.~\cite{wgan} derive an alternate formulation to measure the Wasserstein distance using Kantorovich-Rubinstein duality as shown in Eqn.~\ref{eq:wgan2}.

\begin{align}\label{eq:wgan2}
W\left(\mathbb{P}_{T}, \mathbb{P}_{G}\right)=\sup _{\|f\|_{L} \leq 1} \mathbb{E}_{x \sim \mathbb{P}_{T}}[f(x)]-\mathbb{E}_{x \sim \mathbb{P}_{G}}[f(x)]
\end{align}

Here the supremum is taken over all $1$-Lipschitz continuous functions $f$. To find a function $f$ that satisfies Eqn~\ref{eq:wgan2}, the authors consider a parameterizable \emph{critic function} $D_w: \mathcal{X} \rightarrow \mathbb{R}$. The parameters $w$ of the critic function are chosen by solving the optimization problem shown in Eqn~\ref{eq:wgan3}.

\begin{align}\label{eq:wgan3}
W\left(\mathbb{P}_{T}, \mathbb{P}_{G}\right)= \max_w \mathbb{E}_{x \sim \mathbb{P}_T}\left[D_w(x)\right]-\mathbb{E}_{z \sim \mathcal{N}(0,I)}\left[D_{w}\left(G(z)\right)\right]
\end{align}

Thus, in order to generate realistic synthetic examples that are close to the target distribution, $G$ is trained to minimize the estimate of Wasserstein Distance in Eqn.~\ref{eq:wgan3}. This can be achieved by maximizing the value of the critic function for the generator's examples using the loss function shown in Eqn~\ref{eq:wgan4}. $D$ is trained to solve the optimization problem in Eqn.~\ref{eq:wgan3} by using the loss function in Eqn~\ref{eq:wgan5}.

\begin{align}
z\sim&\mathcal{N}(0,I);\ x \sim \{x_i\}_{i=1}^n \nonumber\\
\mathcal{L}_G &= - D_w(G(z)) \label{eq:wgan4}\\
\mathcal{L}_D &= D_w(G(z))-D_w(x)\label{eq:wgan5}
\end{align}

To ensure that $D$ is $K$-Lipschitz continuous, the original WGAN paper proposed weight clipping. A later work by Gulrajani et al.~\cite{improved} proposed using gradient penalty to improve the stability of training, which we adopt in our proposal.

\section{Sensitivity Studies}\label{app:sensitivity}
In this section we perform sensitivity studies to understand the impact of various attack parameters of MAZE like the query budget ($Q$), number of directions used in gradient estimation ($m$) on the clone accuracy obtained by our attack. In addition, we also quantify the importance of the experience replay step used in our algorithm and the impact of gradient estimation error arising from using zeroth order gradient estimation to approximate the gradient.

\subsection{Sensitivity to Query Budget}

\begin{figure}[htb]
	\centering
    \centerline{\epsfig{file=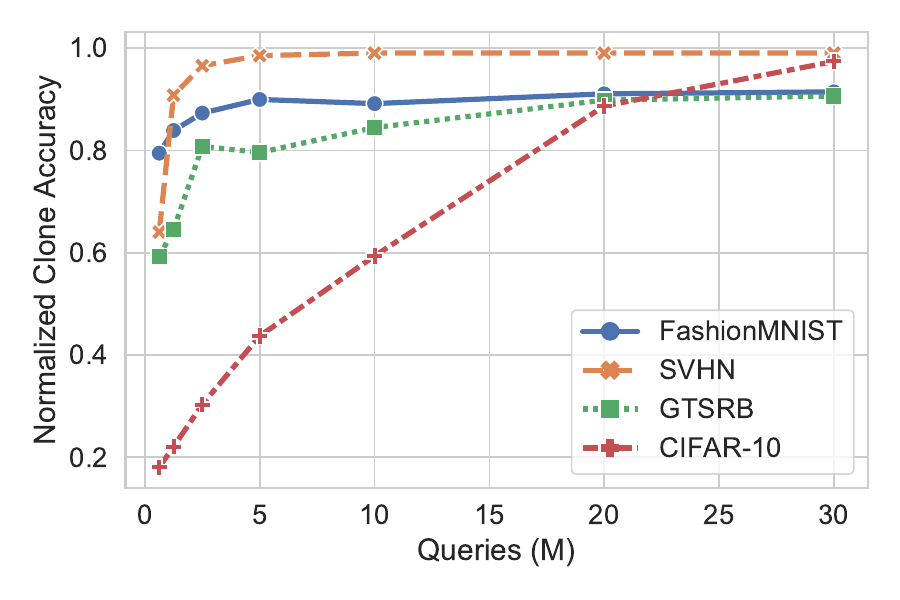, width=0.8\columnwidth}}

	\caption{Normalized clone accuracy versus Query budget ($Q$) for various target models under MAZE attack. Clone accuracy improves with a higher query budget.}

    \label{fig:query_sensitivity}
\end{figure}

A larger query budget allows the attacker to carry out more attack iterations and train a better clone model. To understand the dependence of query budget $Q$ to clone accuracy, we perform sensitivity studies by carrying out our attack for seven different values of $Q$ for each dataset. We set $Q\in\{0.625M,1.25M,2.5M,5M,10M,20M,30M\}$ and report the normalized clone accuracy for each dataset in Fig ~\ref{fig:query_sensitivity}. As expected, we find that the clone accuracy increases with an increase in $Q$. Additionally, we note that the number of queries necessary to reach a given clone accuracy seems to scale with the dimensionality of the input and the difficulty of the target classification task. \ignore{While we use a large query budget of $Q=30M$ to report the clone accuracy in Table~\ref{table:results}, not all datasets require such a high query budget to reach the desired level of accuracy.}For example, we only require around $5M$ queries to reach a normalized accuracy of $0.95\times$ for SVHN, whereas for CIFAR-10, which is a harder dataset to classify, we require around $30M$ queries to reach the same accuracy.

\subsection{Sensitivity to Number of Directions for Estimating the Gradient}

\begin{figure}[htb]
	\centering
    \centerline{\epsfig{file=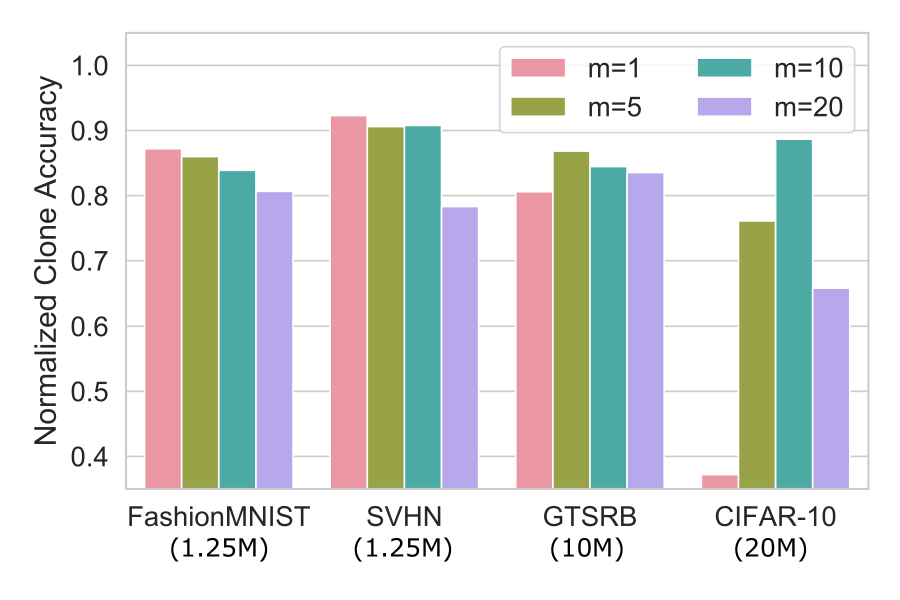, width=0.8\columnwidth}}

	\caption{Normalized clone accuracy versus number of gradient directions ($m$) used in MAZE. Increasing $m$ results in a lower gradient estimation error, but consumes more queries to update $G$, leaving fewer queries to train the clone model.}
    \label{fig:m_sensitivity}
\end{figure}

Our attack uses a generative model to synthesize the queries necessary to perform model stealing. The loss function of the generator $\mathcal{L}_{G}$ involves the evaluation of a black-box model $T$. As gradient information is unavailable, our attack uses zeroth-order gradient estimation instead to approximate the gradient of the loss function $\hat{\nabla}_{\theta_{G}} \mathcal{L}_{G}$ required to update the generator parameters $\theta_G$. The estimation error of this zeroth-order approximation is inversely related to the number of gradient estimation directions ($m$) used in our numerical approximation of the gradient (Eqn.~\ref{eq:update_g}). By using a larger value of $m$, we can get a more accurate estimate of the gradient. Unfortunately, increasing $m$ also increases the queries that need to be made to the target model as described by Eqn.~\ref{eq:query_cost}. This means that given a fixed query budget $Q$, increasing $m$ results in more queries being consumed to update $G$, leaving fewer queries to train the clone model. To understand this trade-off, we fix the query budget and perform our attack by setting $m$ to four different values ($m\in\{1,5,10,20\}$). With a large query budget of $30M$, changing $m$ has limited impact as most clone models achieve high accuracy regardless of the value used for $m$. Hence, for this analysis, we use smaller query budgets instead to magnify the impact of $m$ on the clone accuracy. We set the query budget $Q$ to $1.25M$ for FashionMNIST and SVHN, $10M$ for GTSRB, and $20M$ for CIFAR-10.  The normalized accuracy of clones obtained for these different values of $m$ on various datasets are shown in Fig.~\ref{fig:m_sensitivity}.  

Our results show that for FashionMNIST and SVHN, lowering the value of $m$ results in an improvement in clone accuracy. Setting $m$ to a smaller value allows more queries to be used to train the clone model, while sacrificing the accuracy of the  gradient update of the generator. This seems to be a favorable trade-off for simpler datasets. Similarly, for GTSRB we find that reducing the number of directions from $m=20$  to $m=5$ improves clone accuracy. However, reducing it further to $m=1$ leads to a degradation in the clone accuracy. We observe a similar trend for CIFAR-10 as well. While reducing $m$ from $20$ to $10$ improves accuracy, reducing it further causes a degradation. This degradation in accuracy by reducing $m$ for GTSRB and CIFAR-10 can be attributed to the increased error in gradient estimation with fewer gradient estimation directions. Thus, in a query limited setting, varying $m$ provides a trade-off between the number of queries used to update the generator $G$ and clone model $C$. The optimal value of $m$ for  model stealing depends on the complexity of the target dataset being attacked.



\begin{figure}[htb]
	\centering
    \centerline{\epsfig{file=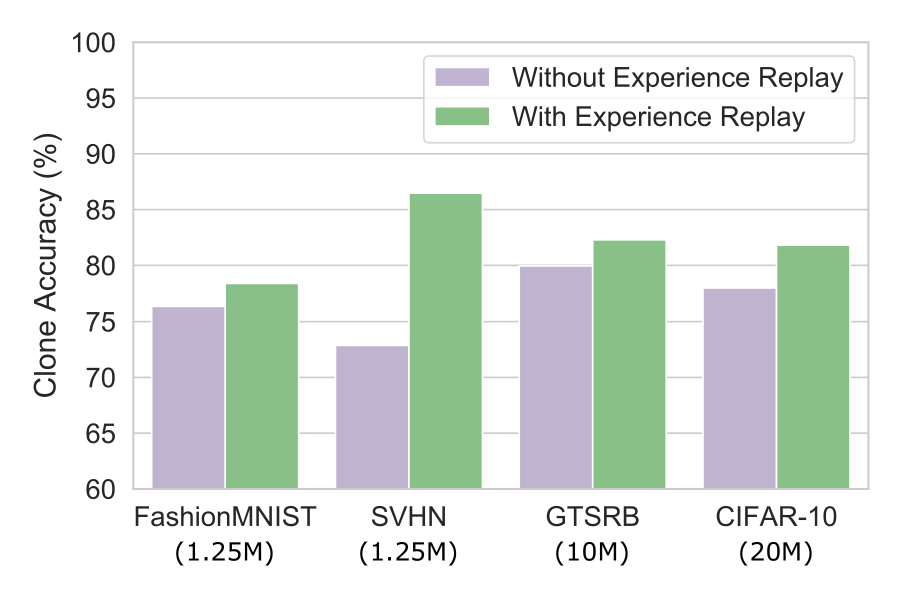, width=0.8\columnwidth}}

	\caption{Impact of \emph{Experience Replay} on MAZE. On average, \emph{Experience Replay} improves clone accuracy by $7.3\%$.}
    \label{fig:exp_replay_sensitivity}
\end{figure}

\subsection{Importance of Experience Replay}
For training, MAZE uses \emph{experience replay}, in which the clone model is retrained on previously seen examples throughout the course of the attack. This is necessary to avoid \emph{catastrophic forgetting}, wherein the clone model performs poorly on examples from the earlier part of the training process. Additionally, it also ensures that the generator does not produce redundant examples that are similar to the ones seen in the earlier part of the training. To understand the importance of \emph{experience replay}, we carry out two versions of the attack: one with experience replay and the other without, and compare the accuracy of the respective clone models. The results of this study are shown in Fig. ~\ref{fig:exp_replay_sensitivity}. Our results show that on average, \emph{experience replay} improves the accuracy of the clone model by $7.3\%$. Thus, experience replay is an important component of MAZE. 

\subsection{Impact of Error in Gradient Estimation}

MAZE uses numerical methods to approximate the gradient through the black-box target model $T$ to update the parameters of the generator model. To understand how the gradient estimation error impacts the accuracy of the clone model, we repeat MAZE by assuming that we have access to perfect gradient information. For the purpose of analysis in this section only, we obtained perfect gradient information by treating the target $T$ as a white-box model that allows back-propagation in the Generator Training phase shown in Algorithm~\ref{alg:attack}. By comparing the results of our attack with the accuracy of the clone model trained with perfect gradient information, we can understand how the gradient estimation error impacts the accuracy of the clone model obtained from our attack.

\begin{figure}[htb]
	\centering

    \centerline{\epsfig{file=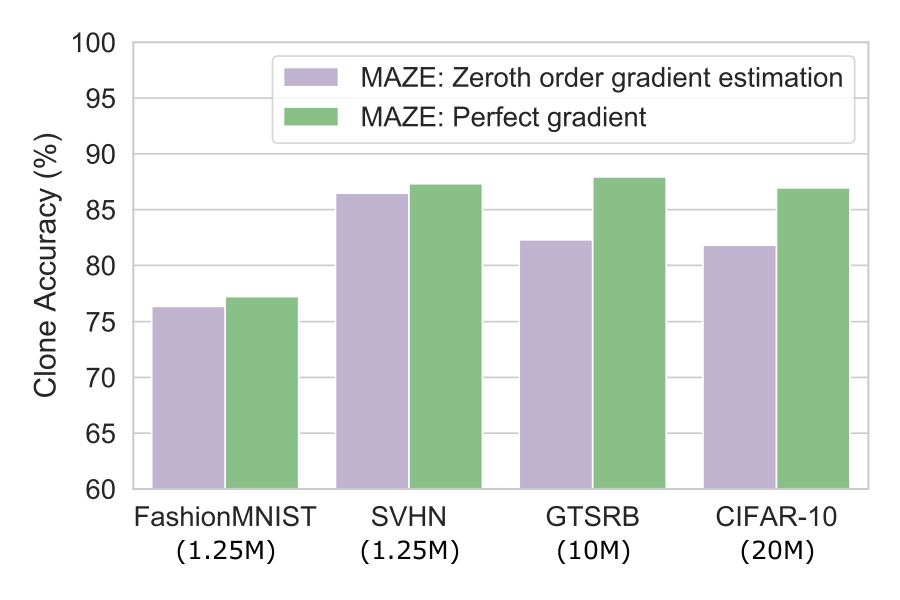, width=0.8\columnwidth}}

    \caption{Impact of gradient estimation error by comparing clone accuracy of MAZE and MAZE with perfect gradient information. On average, MAZE suffers a 3.8\% accuracy loss due to error in estimating the gradient.}
    \label{fig:abm_vs_maze}

\end{figure}

Figure~\ref{fig:abm_vs_maze} shows the clone accuracy of MAZE with zeroth-order gradient estimation and MAZE with perfect gradient information.  We use a reduced query budget of $1.25M$ for FashionMNIST and SVHN, $10M$ for GTSRB, and $20M$ for CIFAR-10.  We observe that there is a slight improvement in clone accuracy (on average, 3.8\%) when perfect gradient information is available.  This shows that our approximation of gradient is reasonably accurate and therefore MAZE is able to train clone model with high accuracy.  

When the query budget is increased, the error in estimating the gradient is tolerated by the training algorithm, and we observe that at a budget of 30 million queries the difference in clone accuracy with estimated gradient and perfect gradient is negligible.




\section{MAZE-PD: Impact on Query Budget} \label{app:maze_pd_query_budget}

To understand the reduction in query budget with MAZE-PD, we compare the query budget necessary to reach a minimum normalized clone accuracy of $0.90\times$ between MAZE-PD and MAZE in Table~\ref{table:maze_vs_mazepd}. Our results show that MAZE-PD offers a reduction of $2\times$ to $24\times$ in the query budget compared to MAZE. While we see a considerable reduction in query budget with just 100 examples, we expect this to reduce further with more training examples.

\begin{table}[htb]
\caption{Comparison of query budgets needed to reach a normalized clone accuracy of 0.90$\times$ with MAZE-PD and MAZE. MAZE-PD reduces the query budget by up to $24\times$.}
\begin{tabular}{|l||c|c|c|}
\hline
\textbf{Target   Models} & \textbf{MAZE}    & \textbf{MAZE-PD} & \textbf{Reduction} \\\hline \hline
FashionMNIST             & \textgreater{}30 M & 2.5 M              & \textgreater{}12$\times$                    \\ \hline
SVHN                     & 1.25 M            & 0.675 M            & 2$\times$                     \\ \hline
GTSRB                    & 30 M              & 1.25 M            & 24$\times$                    \\ \hline
CIFAR-10                 & 30 M              & 10 M              & 3$\times$                     \\ \hline
\end{tabular}
\label{table:maze_vs_mazepd}
\end{table}

\begin{figure*}[tb]
	\centering
    \centerline{\epsfig{file=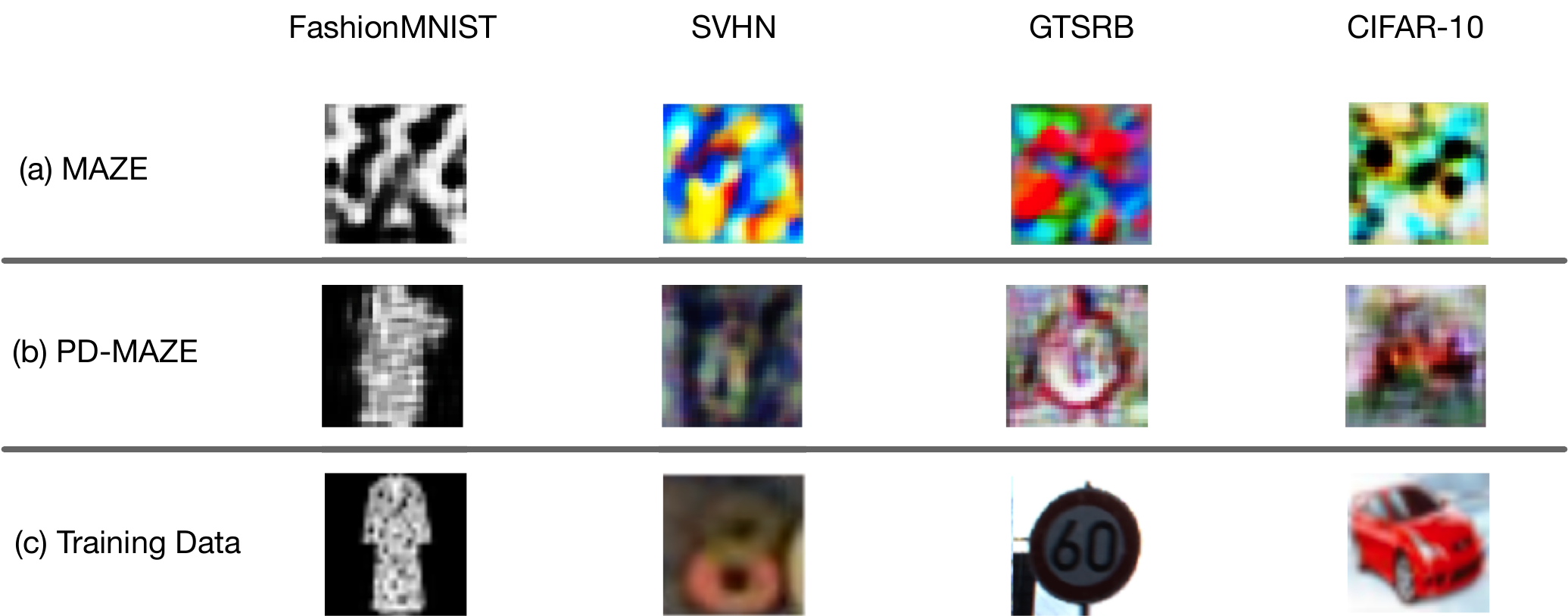, width=0.85\textwidth}}
    \vspace{0.15 in}
	\caption{Comparing images produced by the generative models for (a) MAZE and (b) MAZE-PD with (c) actual training data. MAZE-PD produces images that are visually similar to the target distribution by including the WGAN objective to train the generative model, leading to an improvement in query efficiency over MAZE.}
    \label{fig:generator_images}
\end{figure*}

\section{Comparing Synthetic Images Generated by MAZE and MAZE-PD}\label{app:synthetic_images}

Both MAZE and MAZE-PD use a generator to create synthetic images that are used to query the target model.  MAZE produces these images without relying on any input data from the target model.  On the other hand, MAZE-PD aims to use the available data to encourage the distribution of images produces by the generator $\mathbb{P}_G$ to be closer to the target distribution $\mathbb{P}_T$ using the WGAN loss to train the generator. Thus, MAZE-PD is expected to produce visually closer images to the distribution of the target model, and this is the reason for the increased accuracy and speed of MAZE-PD.


To better understand the quality of the images created by MAZE-PD we compare a small number of representative images that are produced by MAZE-PD with MAZE.  Fig~\ref{fig:generator_images} shows four images produced by MAZE and MAZE-PD. We also show four representative images from the corresponding dataset. It can be seen that the images produced by MAZE-PD are visually similar to the images from the target distribution. For example, for FashionMNIST, the synthetic image produced by MAZE-PD resembles a garment. For GTSRB, it resembles a traffic sign.  For SVHN, it resembles the number "$8$". And, for CIFAR-10, it resembles an automobile. 

Furthermore, we can see a clear distinction between the foreground and the background for the images produced by MAZE-PD. However, such separation of foreground and background is absent in the synthetic images produced by MAZE. Thus, using the WGAN objective in the training of the GAN encourages the generator to produce synthetic images that are closer to the target distribution, likely resulting in MAZE-PD generating clones with higher accuracy using fewer queries.

\section{Work on Data-free Knowledge Distillation}\label{app:related_dfkd}

We discuss two recent works on data-free KD that are closely related to our proposal and also explain why these works cannot be used directly to perform model stealing as they require white-box access to the target model.

\vspace{0.1 in}
\noindent\textbf{Adversarial Belief Matching (ABM)~\cite{abm}:} ABM performs knowledge distillation by using images generated from a generative model $G(z; \phi)$. This generative model is trained to produce inputs $x$ such that the predictions of the teacher model $T(X;\theta_T)$ and the predictions of the student model $S(x;\theta_S)$ are dissimilar. Specifically, $G$ is trained to maximize the KL-divergence between the predictions of the teacher and student model using the loss function shown in  Eqn.~\ref{eq:abm_g}. The student model, on the other hand, is trained to match the predictions of the teacher by minimizing the KL-divergence between the predictions of $T$ and $S$ using the loss function in Eqn. ~\ref{eq:abm_s}. By iteratively updating the generator and student model, ABM performs knowledge distillation between $T$ and $S$.

\begin{align} 
x &= G(z) \\
\mathcal{L}_G &= - D_{K L}\left(T\left(\boldsymbol{x}\right) \|\ S\left(\boldsymbol{x}\right)\right) \label{eq:abm_g}\\
\mathcal{L}_S &= D_{K L}\left(T\left(\boldsymbol{x}\right) \|\ S\left(\boldsymbol{x}\right)\right)\label{eq:abm_s}
\end{align}
\vspace{0.1 in}

In addition to the basic idea presented above, ABM also uses an additional {\em Attention Transfer (AT)}~\cite{at} term in the loss function of the student. AT tries to match the attention maps of the intermediate activations between the student and teacher networks. 

The training process of the generator model in ABM assumes white-box access to the target model as the loss function of $G$ (Eqn.~\ref{eq:abm_g}) requires backpropagating through the target model $T$. Moreover, ABM also uses AT, which requires access to the intermediate activations of $T$. Due to these requirements, ABM cannot be directly used in the black-box setting of model stealing attacks.

\vspace{0.05 in}
\noindent\textbf{Dreaming To Distill (DTD)~\cite{dreaming}:} Given a DNN model $T$ and a target class $y$, DTD proposes to use the loss function of $T$ along with various image-prior terms to generate realistic images that resemble the images from the target class in the training distribution of $T$ by framing it as an optimization problem as shown in Eqn.~\ref{eq:dream_prior}. 

\begin{align}
\min _{x} \mathcal{L}(T(x), y))+\mathcal{R}_p(x)\label{eq:dream_prior}
\end{align}

In this equation, $\mathcal{R}_p$ is the image prior regularization term which steers the synthetic image $x$ away from unnatural images. The key insight of DTD is that the statistics of the batch norm layers (channel-wise mean and variance information of the activation maps for the training data) can be used to construct an image prior term that helps craft realistic inputs by performing the optimization in Eqn.~\ref{eq:dream_prior} . In the knowledge distillation setting, the inputs generated using this technique can be used to train a student model $S$ from the predictions of the teacher model $T$. 

 Note that, much like ABM, DTD also requires white-box access to the target model since the optimization problem in Eqn.~\ref{eq:dream_prior} requires backpropagating through the teacher model. Furthermore, this work requires us to know the running mean and variance information stored in the batch norm layer, which is unavailable in the black-box setting of the model stealing attack.

\section{Defending against MAZE}

MAZE is the first data-free model stealing attack that can effectively produce high accuracy clone models for multiple vision-based DNN models. Several recent works have been proposed to defend against model stealing attacks. We discuss the applicability of these defenses against our attack and explain their limitations. 

MAZE requires access to the prediction probabilities of $T$ in order to estimate gradient information. Thus, a natural way to defend against it would be to limit access to the predictions of the model to the end-user by restricting the output of the model only to provide hard-labels. Such a defense would make it harder, although not necessarily impossible, for an adversary to estimate the gradient by using numerical methods~\cite{chen2019boundary,cheng2018query,ilyas2018black}. Unfortunately, such a defense may limit  benign users from using the prediction probabilities from the service for downstream processing tasks. 

Another potential method to defend against MAZE is to prevent an adversary from accessing the true predictions of the model by perturbing the output probabilities with some noise. Several defenses have been proposed along these lines ~\cite{pp,dp,kariyappa2019defending}. Unfortunately, a key shortcoming of perturbation-based defense is that it can destroy information contained in the class probabilities that can be important for a benign user of the service for downstream processing tasks.  Furthermore, such a defense  can reduce classification accuracy for benign users, which is undesirable.

Yet another way to reduce the effectiveness of our attack may be to limit the number of queries that each user can make to the service. However, an adversary could circumvent such a defense by launching a distributed attack where the task of attacking the model is split across multiple users. Furthermore, limiting the number of queries may also constrain some of the legitimate users of the service from making benign queries, which may also be undesirable.


\end{document}